\documentclass[runningheads]{llncs}
\usepackage{graphicx}
\usepackage{amsmath}
\usepackage[toc,page]{appendix}
\usepackage{bbm} 
\usepackage{pdfpages} 
\usepackage{amssymb} 
\usepackage{multirow} 
\usepackage{bm} 
\usepackage{booktabs}
\usepackage{color}
\usepackage{array}
\begin{document}

\title{Explaining predictive models with mixed features using Shapley values and conditional inference trees}
\titlerunning{Explaining models with Shapley values and ctree}
%
\author{Annabelle Redelmeier \and
	Martin Jullum \and Kjersti Aas}
\authorrunning{Redelmeier et al.}
%

\institute{Norwegian Computing Center, P.O. Box 114, Blindern, N-0314 Oslo, Norway\\
	\email{\{annabelle.redelmeier,jullum,kjersti\}@nr.no}}
\maketitle 
\begin{abstract}
It is becoming increasingly important to explain complex, black-box machine learning models. Although there is an expanding literature on this topic, Shapley values stand out as a sound method to explain predictions from any type of machine learning model. The original development of Shapley values for prediction explanation relied on the assumption that the features being described were independent. This methodology was then extended to explain dependent features with an underlying continuous distribution. In this paper, we propose a method to explain mixed (i.e. continuous, discrete, ordinal, and categorical) dependent features by modeling the dependence structure of the features using conditional inference trees. We demonstrate our proposed method against the current industry standards in various simulation studies and find that our method often outperforms the other approaches. Finally, we apply our method to a real financial data set used in the 2018 FICO Explainable Machine Learning Challenge and show how our explanations compare to the FICO challenge Recognition Award winning team.  

\end{abstract}
\section{Introduction}\label{sec:intro}
Due to the ongoing data and artificial intelligence (AI) revolution, an increasing number of crucial decisions are being made with complex automated systems. It is therefore becoming ever more important to understand how these systems make decisions. Such systems often consist of `black-box' machine learning models which are trained to predict an outcome/decision based on various input data (i.e. features). Consider, for instance, a model that predicts the price of car insurance based on the features age and gender of the individual, type of car, time since the car was registered, and number of accidents in the last five years. 
For such a system to work in practice, 
the expert making the model, the insurance brokers communicating the model, and the policyholders vetting the model should know which features drive the price of insurance up or down.
 
Although there are numerous ways to explain complex models, one way is to show how the individual features contribute to the overall predicted value for a given individual\footnote{Here, `individual' could be an individual person or an individual non-training observation - not necessarily a person.}.
The Shapley value framework is recent methodology to calculate these contributions \cite{Lundberg,Strumbelj,Strumbelj2}. 
In the framework, a Shapley value is derived for each feature given a prediction (or `black-box') model and the set of feature values for the given individual. 
The methodology is such that the sum of the Shapley values for the individual equals their prediction value so that the features with the largest (absolute) Shapley values are the most important.

The Shapley value concept is based on economic game theory. The original setting is as follows: Imagine a game where $N$ players cooperate in order to maximize the \textit{total gains} of the game. Suppose further that each player is to be given a certain payout for his or her efforts. 
Lloyd Shapley \cite{Shapley53} discovered a way to distribute the total gains of the game among the players according to certain desirable axioms. For example, players that do not contribute anything get a payout of 0; two players that contribute the same regardless of other players get the same payout; and the sum of the payouts equals the total gains of the game. 
A player's payout is known as his or her \textit{Shapley value}.

\cite{Lundberg,Strumbelj,Strumbelj2} translate Shapley values from the game theory setting to a machine learning setting.
The cooperative game becomes the individual, the total gains of the game become the prediction value, and the players become the feature values. Then, analogous to game theory, the Shapley value of one of the features (called the \textit{Shapley value explanation}) is how the feature contributes to the overall prediction value. 

\begin{figure}[ht]
	\centering
	\includegraphics[width = 12cm, height = 5.5cm]{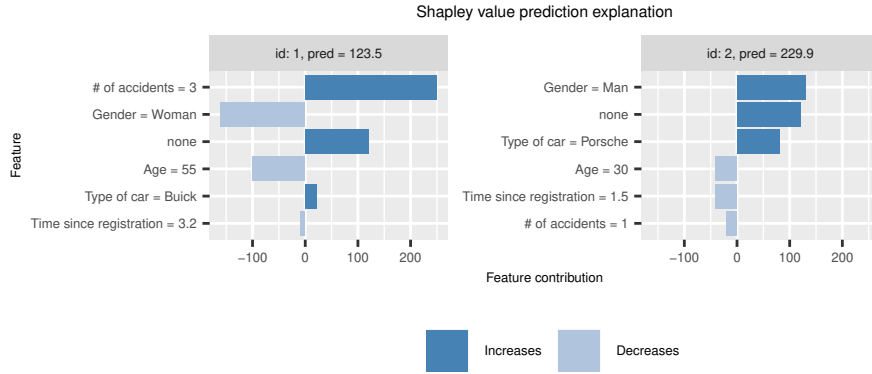}
	\caption{An example of using Shapley values to show how the predicted price of car insurance can be broken down into the respective features.}
	\label{fig:example_carinsurance}
\end{figure}

Figure \ref{fig:example_carinsurance} shows how such Shapley value explanations can be visualized for two examples of the aforementioned car insurance scenario.
For the individual on the left, `number of accidents' pulls the predicted insurance price up (its Shapley value is positive) whereas `gender' and `age' pull it down. The features `type of car' and `time since registration' only minimally affect the prediction. 
For the individual on the right, `gender' and `type of car' pull the predicted insurance price up whereas `age', `time since registration', and `number of accidents' pull it marginally down. The sum of the Shapley values of each individual gives the predicted price of insurance (123.5 and 229.9 USD/month, respectively). 
Note that `none' is a fixed average prediction contribution not due to any of the features in the model.

As we demonstrate in Section \ref{sec:Shapley}, calculating Shapley values is not necessarily straightforward or computationally simple. 
To simplify the estimation problem, \cite{Lundberg,Strumbelj,Strumbelj2} assume the features are independent. However, \cite{aas2019Explaining} shows that this may lead to severely inaccurate Shapley value estimates and, therefore, incorrect explanations when the features are not independent. 
\cite{aas2019Explaining} extends \cite{Lundberg}'s methodology to handle dependent \textit{continuously} distributed features 
by modeling/estimating the dependence between the features.
However, as exemplified by the aforementioned car insurance example, practical modeling scenarios often involve a mixture of different feature types: continuous (age and time since the car was registered), discrete (number of accidents in the last five years), and categorical (gender and type of car). Thus, there is a clear need to extend the Shapley value methodology to handle dependent mixed (i.e. continuous, discrete, ordinal, categorical) features. 

While it is, in principle, possible to naively apply some of the methods proposed by \cite{aas2019Explaining} to discrete or categorical features, it is unlikely that they will function well. This will typically require encoding categorical features with $L$ different categories into $L-1$ new indicator features using one-hot encoding. The main drawback of this approach is that the feature dimension increases substantially unless there are very few categories.
Computational power is already a non-trivial issue with the Shapley value framework (see \cite{aas2019Explaining}), so this is not a feasible approach unless the number of categories or features is very small.

The aim of this paper is to show how we can extend the Shapley value framework to handle mixed features without assuming the features are independent. 
We propose to use a special type of tree model, known as conditional inference trees \cite{hothorn2006unbiased}, to model the dependence between the features. This is similar to \cite{aas2019Explaining}'s extension of \cite{Lundberg}'s work but for mixed features.
We use tree models since they are inherently good at modeling both simple and complex dependence structures in mixed data types \cite{elith2008working}. The conditional inference tree model has the additional advantage of naturally extending to multivariate responses, which is required in this setting. Since conditional inference trees handle categorical data, this approach does not require one-hot encoding any features resulting in a much shorter computation time. In addition, we do not run the risk of estimating one-hot encoded features using a method not designed for the sort.

The rest of the paper is organized as follows. We begin by explaining the fundamentals of the Shapley value framework in an explanation setting in Section \ref{sec:Shapley} and then outline how to extend the method to mixed features using conditional inference trees in Section \ref{sec:ctree}. In Section \ref{sec:simulation}, we present various simulation studies for both continuous and categorical features that demonstrate that our method works in a variety of settings. Finally, in Section \ref{sec:data}, we apply our method to the 2018 FICO Explainable Machine Learning Challenge data set and show how the estimated Shapley values differ when calculated using various feature distribution assumptions. 
We also compare the feature importance rankings calculated using Shapley values with the rankings calculated by the 2018 FICO challenge Recognition Award winning team. 
In Section \ref{sec:conclusion}, we conclude.

The Shapley methodology from \cite{aas2019Explaining} is implemented in the software package \verb|shapr| \cite{Sellereite2020} in the \verb|R| programming language \cite{citing_R_}. 
Our new approach is implemented as an extension to the \verb|shapr| package.
To construct the conditional inference trees in \verb|R|, we use the packages \verb|party| and \verb|partykit| \cite{hothorn2006unbiased,partykit_package}. 

\section{Shapley values}\label{sec:Shapley}

\subsection{Shapley values in game theory}\label{subsec:Shapley_framework}
Suppose we are in a cooperative game setting with $M$ players, $j = 1, \dots, M$, trying to maximize a payoff. Let $\mathcal{M}$ be the set of all players and $\mathcal{S}$ any subset of $\mathcal{M}$. Then the Shapley value \cite{Shapley53} for the $j$th player is defined as 


\begin{equation}\label{eq:shapley}
\phi_j = \sum_{\mathcal{S} \subseteq \mathcal{M} \backslash \{j\}} \frac{|\mathcal{S}|! (M - |\mathcal{S}| - 1)!}{M!} (v(\mathcal{S} \cup \{ j\}) - v(\mathcal{S})).
\end{equation}
$v(\mathcal{S})$ is the contribution function which maps subsets of players to real numbers representing the worth or contribution of the group $\mathcal{S}$ and $|\mathcal{S}|$ is the number of players in subset $\mathcal{S}$. 

In the game theory sense, each player receives $\phi_j$ as their payout.  
From the formula, we see that this payout is just a weighted sum of the player's marginal contributions to each group $\mathcal{S}$. 
Lloyd Shapley \cite{Shapley53} proved that distributing the total gains of the game in this way is `fair' in the sense that it obeys certain important axioms. 

\subsection{Shapley values for explainability}\label{subsec:Shapley_explain}
In a machine learning setting, imagine a scenario where we fit $M$ features, $\bm{x} = (x_1, \dots, x_M)$, to a univariate response $y$ with the model $f(\bm{x})$ and want to explain the prediction $f(\bm{x})$ for a specific feature vector $\bm{x} = \bm{x}^*$. 
\cite{Lundberg,Strumbelj,Strumbelj2} suggest doing this with Shapley values where the predictive model replaces the cooperative game and the features replace the players. To use \eqref{eq:shapley}, \cite{Lundberg} defines the contribution function $v(\mathcal{S})$ as the following expected prediction
\begin{align}\label{eq:cond_expec}
v(\mathcal{S}) = \mathbb{E}[f(\bm{x}) | \bm{x}_\mathcal{S} = \bm{x}^*_\mathcal{S}]. 
\end{align} 
Here $\bm{x}_\mathcal{S}$ denotes the features in subset $\mathcal{S}$ and $\bm{x}_\mathcal{S}^*$ is the subset $\mathcal{S}$ of the feature vector $\bm{x}^*$ that we want to explain. Thus, $v(\mathcal{S})$ denotes the expected prediction given that the features in subset $\mathcal{S}$ take the value $\bm{x}_\mathcal{S}^*$.

Calculating the Shapley value for a given feature $x_j$ thus becomes the arduous task of computing \eqref{eq:shapley} but replacing $v(\mathcal{S})$ with the conditional expectation \eqref{eq:cond_expec}. 
It is clear that the sum in \eqref{eq:shapley} grows exponentially as the number of features, $M$, increases. 
\cite{Lundberg} cleverly approximates this weighted sum in a method they call Kernel SHAP. Specifically, they define the Shapley values as the optimal solution to a certain weighted least squares problem. They prove that the Shapley values can explicitly be written as 
\begin{align}\label{eq:WLS}
\phi = (\bm{Z}^T\bm{W}\bm{Z})^{-1} \bm{Z}^T\bm{W}\bm{v},
\end{align} 
where $\bm{Z}$ is the $2^M \times (M+1)$ binary matrix representing all possible combinations of the $M$ features, $\bm{W}$ is the $2^M \times 2^M$ diagonal matrix containing Shapley weights, and $\bm{v}$ is the vector containing $v(\mathcal{S})$ for every $\mathcal{S}$. The full derivation is described in \cite{aas2019Explaining}. 

To calculate \eqref{eq:WLS}, we still need to compute the contribution function $v(\mathcal{S})$ for different subsets of features, $\mathcal{S}$. When the features are continuous, we can write the conditional expectation \eqref{eq:cond_expec} as
\begin{equation}\label{eq:cond_integral}
\begin{split}
\mathbb{E}[f(\bm{x}) | \bm{x}_\mathcal{S} = \bm{x}_\mathcal{S}^*] &= \mathbb{E}[f(\bm{x}_{\bar{\mathcal{S}}}, \bm{x}_{\mathcal{S}})| \bm{x}_\mathcal{S} = \bm{x}_\mathcal{S}^*] = \int f(\bm{x}_{\bar{\mathcal{S}}}, \bm{x}_{\mathcal{S}}^*) p(\bm{x}_{\bar{\mathcal{S}}} | \bm{x}_{\mathcal{S}} = \bm{x}_{\mathcal{S}}^*) \,d\bm{x}_{\bar{\mathcal{S}}}^*,
\end{split}
\end{equation} where $\bm{x}_{\bar{\mathcal{S}}}$ is the vector of features not in $\mathcal{S}$ and $p(\bm{x}_{\bar{\mathcal{S}}} | \bm{x}_{\mathcal{S}} = \bm{x}_{\mathcal{S}}^*)$ is the conditional distribution of $\bm{x}_{\bar{\mathcal{S}}}$ given $\bm{x}_{\mathcal{S}} = \bm{x}_{\mathcal{S}}^*$. Note that in the rest of the paper we use $p(\cdot)$ to refer to both probability mass functions and density functions (made clear by the context). We also use lower case $x$-s for both random variables and realizations to keep the notation concise.
 
Since the conditional probability function is rarely known, \cite{Lundberg} replaces it with the simple (unconditional) probability function
\begin{equation}\label{eq:cond_is_marg}
p(\bm{x}_{\bar{\mathcal{S}}} | \bm{x}_{\mathcal{S}} = \bm{x}_{\mathcal{S}}^*) = p(\bm{x}_{\bar{\mathcal{S}}}).
\end{equation}
The integral then becomes
\begin{align}\label{eq:expectation_without_cond}
\mathbb{E}[f(\bm{x}) | \bm{x}_\mathcal{S} = \bm{x}_\mathcal{S}^*] = \int f(\bm{x}_{\bar{\mathcal{S}}}, \bm{x}_{\mathcal{S}}^*) p(\bm{x}_{\bar{\mathcal{S}}}) \,d\bm{x}_{\bar{\mathcal{S}}}^*,
\end{align}
which is estimated by randomly drawing $K$ times from the full training data set and calculating
\begin{align}\label{eq:montecarl}
v_{\text{KerSHAP}}(\mathcal{S}) = \frac{1}{K} \sum_{k=1}^K f(\bm{x}_{\bar{\mathcal{S}}}^k, \bm{x}_\mathcal{S}^*), 
\end{align} where $\bm{x}_{\bar{\mathcal{S}}}^k$, $k = 1, \dots, K$ are the samples from the training set and $f(\mathbf{\cdot})$ is the estimated prediction model.

Unfortunately, when the features are not independent, \cite{aas2019Explaining} demonstrates that naively replacing the conditional probability function with the unconditional one leads to very inaccurate Shapley values. \cite{aas2019Explaining} then proposes multiple methods for estimating $p(\bm{x}_{\bar{\mathcal{S}}} | \bm{x}_{\mathcal{S}} = \bm{x}_{\mathcal{S}}^*)$ without relying on the naive assumption in \eqref{eq:cond_is_marg}.
However, these methods are only constructed for continuous features. In the next section we demonstrate how we can use conditional inference trees to extend the current Shapley framework to handle mixed features. 

\section{Extending the Shapley framework with conditional inference trees}\label{sec:ctree}

Conditional inference trees (ctree) \cite{hothorn2006unbiased} is a type of recursive partitioning algorithm like CART (classification and regression trees) \cite{breiman1984classification} and C4.5 \cite{quinlan2014c4}. Just like these algorithms, ctree builds trees recursively, making binary splits on the feature space until a given stopping criterion is fulfilled. 
The difference between ctree and CART/C4.5 is how the feature/split point and stopping criterion are chosen. CART and C4.5 solve for the feature and split point simultaneously: each feature and split point is tried together and the best pair is the combination that results in the smallest error (often based on the squared error loss or binary cross-entropy loss depending on the response). 
Ctree, on the other hand, proceeds sequentially: the splitting feature is chosen using statistical significance tests and then the split point is chosen using any type of splitting criterion \cite{hothorn2006unbiased}. According to \cite{hothorn2006unbiased}, choosing the splitting feature without first checking for the potential split points avoids being biased towards features with many split points. 
In addition, unlike CART and C4.5, ctree is defined independently of the dimension of the response variable. This is advantageous since proper handling of multivariate responses is crucial for our problem.

\subsection{Conditional inference tree algorithm}\label{subsec:ctree_alg}

Suppose that we have a training data set with $p$ features, a $q$ dimensional response, and $n$ observations: $\{\bm{y}_i, \bm{x}_i \}_{i = 1, \dots, n}$ with $\bm{y}_i = (y_{i1}, \dots y_{iq})$ and $\bm{x}_i = (x_{i1}, \dots, x_{ip})$. Suppose further that the responses come from a sample space $\mathcal{Y} = \mathcal{Y}_1 \times \dots \times \mathcal{Y}_q$ and the features come from a sample space $\mathcal{X} = \mathcal{X}_1 \times \dots \times \mathcal{X}_p$. 
Then conditional inference trees are built using the following algorithm:

\begin{enumerate}
	\item For a given node in the tree, test the global null hypothesis of independence between all of the $p$ features and the response $\bm{y}$. If the global hypothesis cannot be rejected, do not split the node. Otherwise, select the feature $x_j$ that is the least independent of $\bm{y}$. 
	\item Choose a splitting point in order to split $\mathcal{X}_{j}$ into two disjoint groups. 
\end{enumerate}
Steps 1 and 2 are repeated until no nodes are split. 


The global null hypothesis can be written as 
\begin{align*}
H_0: \cap_{j=1}^p H_0^j,
\end{align*}
where the $p$ partial hypotheses are 
\begin{align*}
H_0^j: F(\bm{Y}|X_j) = F(\bm{Y}),
\end{align*}
and $F(\cdot)$ is the distribution of $\bm{Y}$.

Specifically, we calculate the $p$ $P$-values for the partial hypotheses and combine them to form the global null hypothesis $P$-value. 
If the $P$-value for the global null hypothesis is smaller than some predetermined level $\alpha$, we reject the global null hypothesis and assume that there is some dependence between the features and the response. 
The feature that is the least independent of the response (i.e. has the smallest partial $P$-value) becomes the splitting feature. If the global null hypothesis is not rejected, we do not split the node. 
The size of the tree is controlled using the parameter $\alpha$. 
As $\alpha$ increases, we are more likely to reject the global null hypothesis and therefore split the node. This results in deeper trees.
However, if $\alpha$ is too large, we risk that the tree overfits the data.

Step 2 can be done using any type of splitting criterion, specifically it can be done with the permutation test framework devised by \cite{hothorn2006unbiased}.  
Note that this method is not tied to a specific feature type and can be used with mixtures of continuous, discrete, ordinal, and categorical features.
We refer the reader to the original paper \cite{hothorn2006unbiased} for more details on how to form the test statistic, associated distribution, and $P$-values.

\subsection{Extending the Shapley value framework with conditional inference trees}\label{subsec:shapley_ctree}

As already mentioned, one of the main limitations with the Shapley value framework is estimating the contribution function \eqref{eq:cond_expec} when the conditional distribution of the features is unknown but the features are assumed dependent. 
\cite{aas2019Explaining} estimates \eqref{eq:cond_expec} by modeling the conditional probability density function
\begin{align}\label{eq:conditional_x}
p(\bm{x}_{\bar{\mathcal{S}}} | \bm{x}_{\mathcal{S}} = \bm{x}_{\mathcal{S}}^*)
\end{align}
using various approaches. Then, \cite{aas2019Explaining} samples $K$ times from this modeled conditional distribution function and uses these samples to estimate the integral \eqref{eq:cond_integral} using \eqref{eq:montecarl}.

We extend this approach to mixed features by modeling the conditional distribution function \eqref{eq:conditional_x} using conditional inference trees. 
We fit a tree to our training data where the features are $\bm{x}_{\mathcal{S}}$ and the response is $\bm{x}_{\bar{\mathcal{S}}}$ with the algorithm described in Section \ref{subsec:ctree_alg}. 	
Then for a given $\bm{x}_\mathcal{S}^*$, we find its leaf in the tree and sample $K$ times from the $\bm{x}_{\bar{\mathcal{S}}}$ part of the training observations in that node to obtain $\bm{x}_{\bar{\mathcal{S}}}^k$, $k = 1, \dots, K$. Finally, we use these samples to estimate \eqref{eq:cond_integral} using the approximation \eqref{eq:montecarl}. 

We fit a new tree to every combination of features $\bm{x}_{\mathcal{S}}$ and response  $\bm{x}_{\bar{\mathcal{S}}}$. 
Once $v(\mathcal{S})$ is estimated for every $\mathcal{S}$, we follow \cite{lundberg2018consistent}'s steps and estimate the Shapley value of this feature with \eqref{eq:WLS}.
Since conditional inference trees handle continuous, discrete, ordinal, and categorical features; univariate and multivariate responses; and any type of dependence structure, using conditional inference trees to estimate \eqref{eq:conditional_x} is a natural extension to \cite{aas2019Explaining}'s work.  Below, we use the term \textit{ctree} to refer to estimating Shapley value explanations using conditional inference trees. 

\section{Simulation studies}\label{sec:simulation}

In this section, we discuss two simulation studies designed to compare different ways to estimate Shapley values. Specifically, we compare our ctree estimation approach with \cite{Lundberg}'s independence estimation approach (below called \textit{independence}) and \cite{aas2019Explaining}'s empirical and Gaussian estimation approaches. A short description of each approach is in Table \ref{tbl:sim_study_outline2}.

\begin{table}[ht]
	\centering
	\begin{tabular}{>{\centering}p{0.2\textwidth} >{\centering}p{0.15\textwidth} >{\arraybackslash}p{0.68\textwidth}}
		\textbf{Method} & \textbf{Citation} & \textbf{Description} \\
		\midrule
		independence & \cite{lundberg2018consistent} & Assume the features are independent. Assume \eqref{eq:cond_expec} is \eqref{eq:expectation_without_cond} and estimate it with \eqref{eq:montecarl} where $x_{\bar{\mathcal{S}}}^k$ are sub-samples from the training data set. \\
		\hline
		empirical & \cite{aas2019Explaining} & Calculate the distance between the set of features being explained and every training instance. Use this distance to calculate a weight for each training instance. Approximate \eqref{eq:cond_expec} using a function of these weights. \\
		\hline
		Gaussian (100) & \cite{aas2019Explaining} & Assume the features are jointly Gaussian. Estimate the mean/covariance of this conditional distribution and then sample 100 times from this distribution. Estimate \eqref{eq:cond_expec} with \eqref{eq:montecarl} using this sample.\\
		\hline
		Gaussian (1000) & \cite{aas2019Explaining} & The same as Gaussian (100), but we sample 1000 times.\\
		\hline
		ctree & & See Section \ref{subsec:shapley_ctree}. Set $\alpha = 0.5$. \\
		\hline
		ctree-onehot & & Convert the categorical features into one-hot encoded features and then apply the algorithm in Section \ref{subsec:shapley_ctree} to these binary features. This approach is used only as a reference.\\
		\midrule
	\end{tabular}
	\vspace{0.2cm}
	\caption{A short description of the approaches used to estimate \eqref{eq:conditional_x} in the simulation studies.}
	\label{tbl:sim_study_outline2}
\end{table}

The independence, empirical, and Gaussian approaches are all implemented in the \verb|R| package \verb|shapr| \cite{Sellereite2020}.
We implement the ctree method in the \verb|shapr| package as an additional method. Building the conditional inference trees for each combination of features is done using either the \verb|party| package or \verb|partykit| package in \verb|R| \cite{hothorn2006unbiased,partykit_package}. Although \verb|party| is faster than \verb|partykit|, it sometimes runs into a convergence error related to the underlying linear algebra library in \verb|R| (error code 1 from Lapack routine 'dgesdd'). We therefore fall back to \verb|partykit| when this error occurs. Both packages typically give identical results.

In the first simulation study, we simulate only categorical features and in the second, we simulate both categorical and continuous features. Then, we estimate the Shapley values of each test observation with the methods in Table \ref{tbl:sim_study_outline2} and compare them against the truth using a mean absolute error type of performance measure.

For simplicity, in both situations we restrict ourselves to a linear predictive function of the form
\begin{align}\label{eq:linmod}
f(\bm{x})=
\alpha + \sum_{\{j:j \in \mathcal{C}_{\text{cat}}\}} \sum_{l=2}^L \beta_{jl} \mathbf{1}(x_j=l)  \quad + \sum_{\{j:j \in \mathcal{C}_{\text{cont}}\}} \gamma_j x_j,
\end{align} 
where $\mathcal{C}_{\text{cat}}$ and $\mathcal{C}_{\text{cont}}$ denote, respectively, the set of categorical and continuous features, $L$ is the number of categories for each of the categorical features, and $\mathbf{1}(x_j=l)$ is the indicator function taking the value 1 if $x_j=l$ and 0 otherwise. $\alpha$, $\beta_{jl}$ for $j \in \mathcal{C}_{\text{cat}}$, $l = 2,\ldots,L$, and $\gamma_j$ for $j \in \mathcal{C}_{\text{cont}}$ are the parameters in the linear model. We define $M = |\mathcal{C}_{\text{cat}}| + |\mathcal{C}_{\text{cont}}|$, where $|\mathcal{C}_{\text{cat}}|$ and $|\mathcal{C}_{\text{cont}}|$ denote the number of categorical and continuous features, respectively.

The empirical and Gaussian methods cannot handle categorical features. For these methods, we transform the categorical features into one-hot encoded features. If the categorical feature originally has $L$ categories, the one-hot encoded transformation creates $L-1$ binary features representing the second, third (etc) categories. The first category is represented by the intercept. The Shapley value of each categorical feature is then the sum of the Shapley values of the corresponding one-hot encoded features.

\subsection{Evaluation method}\label{subsec:evaluation}

We measure the performance of each method based on the mean absolute error (MAE), across both the features and sample space. This is defined as 
\begin{align}\label{eq:MAE_cat}
\text{MAE}(\text{method } q) = \frac{1}{M} \sum_{j=1}^{M}\sum_{i=1}^T p(\bm{x}_i) |\phi_{j, \text{true}}(\bm{x}_i) - \phi_{j, q}(\bm{x}_i) |,
\end{align}
where $\phi_{j, q}(\bm{x})$ and $\phi_{j, \text{true}}(\bm{x})$ denote, respectively, the Shapley value estimated with method $q$ and the corresponding true Shapley value for the prediction $f(\bm{x})$.  In addition, $M$ is the number of features and $T$ is the number of test observations. 
For the case with only categorical features, the set $\{\bm{x}_i: i = 1, \dots, T\}$ corresponds to all the unique combinations of features and $p(\bm{x}_i)$ is the probability mass function of $\bm{x}$ evaluated at $\bm{x}_i$\footnote{If there are many categorical features or number of categories, we instead use a subset of the most likely combinations and scale the probabilities such that they sum to 1 over those combinations.}. 
In the case where we have both categorical and numerical features, the set $\{\bm{x}_i: i = 1, \dots, T\}$ is sampled from the distribution of $\bm{x}$ and $p(\bm{x}_i)$ is set to $1/T$ for all $i$. 

\subsection{Simulating dependent categorical features}\label{subsec:sim_cat_data}

To simulate $M$ dependent categorical features with $L$ categories each, 
we first simulate an $M$-dimensional Gaussian random variable with a specified mean $\bm{\mu}$ and covariance $\bm{\Sigma}$ 
\begin{align}\label{eq:gaussvar}
(\tilde{x}_1, \dots, \tilde{x}_M) \sim N_M(\bm{\mu}, \bm{\Sigma}).
\end{align}
We then transform each feature, $\tilde{x}_j$, into a categorical feature, $x_j$, using the following transformation:
\begin{align}\label{eq:cat_trans}
x_j = l, \text{ if } v_l < \tilde{x}_j \le v_{l + 1}, \text{ for } l = 1, \dots, L \text{ and } j = 1, \dots, M,
\end{align}
where $v_1,\ldots,v_{L+1}$ is an increasing, ordered set of cut-off values defining the categories with $v_1 = -\infty$ and $v_{L + 1} = +\infty$. We redo this $n_{\text{train}}$ times to create a training data set of $M$ dependent categorical features. The strength of the dependencies between the categorical features is controlled by the correlations specified in $\bm{\Sigma}$. Note that the actual value of $x_j$ is irrelevant -- the features are treated as non-ordered categorical features.

For the simulation setting in Section \ref{subsec:both_categorical_and_cont} where there are both categorical and continuous features, we first sample $M=|\mathcal{C}_{\text{cat}}|+|\mathcal{C}_{\text{cont}}|$ features using \eqref{eq:gaussvar}. Then we transform the features $\tilde{x}_j$ where $j \in \mathcal{C}_{\text{cat}}$ to categorical ones using \eqref{eq:cat_trans}, and leave the remaining features untouched (i.e letting $x_j = \tilde{x}_j$, when $j \in \mathcal{C}_{\text{cont}}$). This imposes dependence both within and between all feature types. 

\subsection{Calculating the true Shapley values}\label{subsec:calc_true_Shapley}

To evaluate the performance of the different methods with the MAE from Section \ref{subsec:evaluation}, we need to calculate the true Shapley values, $\phi_{j, \text{true}}(\bm{x}^*)$, $j = 1, \dots, M$, for all feature vectors where $\bm{x}^*= \bm{x}_i$, $i = 1, \ldots, T$. This requires the true conditional expectation \eqref{eq:cond_expec} for all feature subsets $\mathcal{S}$. We compute these expectations differently depending on whether the features are all categorical or whether there are both categorical and continuous features. The linearity of the predictive function \eqref{eq:linmod} helps to simplify the computations for the latter case. Since there is no need of it in the former case, we present that case more generally.

When the features are all categorical, the desired conditional expectation can be written as
\[
\mathbb{E}[f(\bm{x}) | \bm{x}_\mathcal{S} = \bm{x}^*_\mathcal{S}] = \sum_{\bm{x}_{\bar{\mathcal{S}}} \in \mathcal{X}_{\bar{\mathcal{S}}}}  f(\bm{x}^*_{\mathcal{S}},\bm{x}_{\bar{\mathcal{S}}}) p(\bm{x}_{\bar{\mathcal{S}}} | \bm{x}_\mathcal{S} = \bm{x}^*_\mathcal{S}),
\]
where $\mathcal{X}_{\bar{\mathcal{S}}}$ denotes the feature space of the feature vector $\bm{x}_{\bar{\mathcal{S}}}$ which contains $|\bar{\mathcal{S}}|^L$ unique feature combinations. Thus, all we need is the conditional probability $p(\bm{x}_{\bar{\mathcal{S}}} | \bm{x}_\mathcal{S} = \bm{x}^*_\mathcal{S})$ for each combination of $\bm{x}_{\bar{\mathcal{S}}} \in \mathcal{X}_{\bar{\mathcal{S}}}$.
Using standard probability theory, this conditional probability can be written as
\[
p(\bm{x}_{\bar{\mathcal{S}}} | \bm{x}_\mathcal{S}) = \frac{p(\bm{x}_{\bar{\mathcal{S}}}, \bm{x}_{\mathcal{S}})}{p(\bm{x}_{\mathcal{S}})},
\] and then evaluated at the desired $\bm{x}^*_\mathcal{S}$.
Since all feature combinations correspond to hyperrectangular subspaces of Gaussian features, we can compute all joint probabilities exactly using the cut-offs $v_1, \dots, v_{L+1}$:
\[
p(x_1 = l_1, \dots, x_M = l_M) =  P(v_{l_1} < \tilde{x}_1 \le v_{l_1 + 1}, \dots, v_{l_M} < \tilde{x}_M \le v_{l_M + 1}), 
\]
for $l_j = 1, \dots, L$, $j = 1, \ldots, M$. Here $p(\cdot)$ denotes the joint probability mass function of $\bm{x}$ while $P(\cdot)$ denotes the joint continuous distribution function of $\tilde{\bm{x}}$. The probability on the right is easy to compute based on the cumulative distribution function of the multivariate Gaussian distribution (we used the \verb|R| package \verb|mvtnorm| \cite{mvtnorm_R}).
The marginal and joint probability functions based on only a subset of the features are computed analogously based on a subset of the full Gaussian distribution, which is also Gaussian.

For the situation where some of the features are categorical and some are continuous, the computation of the conditional expectation is more arduous. However, due to the linearity of the predictive function \eqref{eq:linmod}, the conditional expectation reduces to a linear combination of two types of univariate expectations. Let $\mathcal{S}_{\text{cat}}$ and $\bar{\mathcal{S}}_{\text{cat}}$ refer to, respectively, the $\mathcal{S}$ and $\bar{\mathcal{S}}$ part of the categorical features, $\mathcal{C}_{\text{cat}}$, with analogous sets $\mathcal{S}_{\text{cont}}$ and $\bar{\mathcal{S}}_{\text{cont}}$ for the continuous features. We then write the desired conditional expectation as 
\begin{align}
\mathbb{E}[f(\bm{x}) | \bm{x}_\mathcal{S} = \bm{x}_\mathcal{S}^*] 
&= \mathbb{E}\left[\alpha + \sum_{j \in \mathcal{C}_{\text{cat}}} \sum_{l=2}^L \beta_{jl} \mathbf{1}(x_j=l) + \sum_{j \in \mathcal{C}_{\text{cont}}} \gamma_k x_j \,\bigg|\,\bm{x}_\mathcal{S} = \bm{x}_\mathcal{S}^*\right] \notag \\
&= \alpha + \sum_{j \in \mathcal{C}_{\text{cat}}} \sum_{l=2}^L \beta_{jl} \mathbb{E}[\mathbf{1}(x_j=l)|\bm{x}_\mathcal{S} = \bm{x}_\mathcal{S}^*] + \sum_{j \in \mathcal{C}_{\text{cont}}} \gamma_j \mathbb{E}[x_j|\bm{x}_\mathcal{S} = \bm{x}_\mathcal{S}^*] \notag \\
&= \alpha + \sum_{j \in \bar{\mathcal{S}}_{\text{cat}}} \sum_{l=2}^L \beta_{jl} \mathbb{E}[\mathbf{1}(x_j=l)|\bm{x}_\mathcal{S} = \bm{x}_\mathcal{S}^*] + \sum_{j \in {\mathcal{S}_{\text{cat}}}} \sum_{l=2}^L \beta_{jl} \mathbf{1}(x_j^* = l) \notag \\
&+ \sum_{j \in \bar{\mathcal{S}}_{\text{cont}}} \gamma_j \mathbb{E}[x_j|\bm{x}_\mathcal{S} = \bm{x}_\mathcal{S}^*] + \sum_{j  \in \mathcal{S}_{\text{cont}}} \gamma_j x^*_j. \notag 
\end{align}
Then, we just need expressions for the two conditional expectations: $\mathbb{E}[\mathbf{1}(x_j=l)|\bm{x}_\mathcal{S} = \bm{x}_\mathcal{S}^*]$ for $j \in \bar{\mathcal{S}}_{\text{cat}}$ and  $\mathbb{E}[x_j|\bm{x}_\mathcal{S} = \bm{x}_\mathcal{S}^*]$ for $j \in \bar{\mathcal{S}}_{\text{cont}}$. To calculate them, we use results from \cite{arellano2006unified} on selection (Gaussian) distributions in addition to basic probability theory and numerical integration. 
Specifically, the conditional expectation for the continuous features takes the form
\begin{align}\label{eq:exp.cat}
\mathbb{E}[x_j|\bm{x}_\mathcal{S}] &= 
\int_{-\infty}^{\infty} x g(x) \frac{p(\bm{x}_\mathcal{S}|x_j=x)}{p(\bm{x}_\mathcal{S})} \,dx,
\end{align}
where $g(x)$ denotes the density of the standard normal (Gaussian) distribution,  $p(\bm{x}_\mathcal{S}|x_j=x)$ is the conditional distribution of $\bm{x}_{\mathcal{S}}$ given $x_j$, and $p(\bm{x}_\mathcal{S})$ is the marginal distribution of $\bm{x}_\mathcal{S}$. The latter two are both Gaussian and can be evaluated at the specific vector $\bm{x}_\mathcal{S}^*$ using the  \verb|R| package \verb|mvtnorm| \cite{mvtnorm_R}. Finally, the integral is solved using numerical integration.

For the second expectation, recall that $x_j=l$ corresponds to the original Gaussian variable $\tilde{x}_j$ falling in the interval $(v_{l},v_{l+1}]$. Then, the conditional expectation for the categorical features takes form
\begin{align*}
\mathbb{E}[\mathbf{1}(x_j=l)|\bm{x}_\mathcal{S}] &= P(v_l < \tilde{x}_j \leq v_{l+1}|\bm{x}_{\mathcal{S}})\\
&= \int_{v_l}^{v_{l+1}} g(x)\frac{p(\bm{x}_\mathcal{S} | \tilde{x}_j = x)}{p(\bm{x}_\mathcal{S})} \,dx,
\end{align*}
which can be evaluated similarly to \eqref{eq:exp.cat} and solved with numerical integration.
Once we have computed the necessary conditional expectations for each of the $2^M$ feature subsets $\mathcal{S}$, we compute the Shapley values using \eqref{eq:WLS}. This goes for both the pure categorical case and the case with both categorical and continuous features.

\subsection{Simulation study with only categorical features}\label{subsec:only_categorical}

We evaluate the performance of the different Shapley value approximation methods in the case of only categorical features with six different experimental setups.
Table \ref{tbl:sim_study_outline} describes these different experiments. 

In each experiment, we sample $n_{\text{train}} = 1000$ training observations using the approach from Section \ref{subsec:sim_cat_data}, where the mean $\bm{\mu}$ is $\bm{0}$ and the covariance matrix is constructed with $\Sigma_{j,j} = 1$, $j = 1, \dots, M$, $\Sigma_{i,j}=\rho$, for $i \neq j$, where $\rho \in \{0, 0.1, 0.3, 0.5, 0.8, 0.9 \}$. We set the response to
\begin{align}\label{eq:response_mod}
y_i = \alpha + \sum_{j=1}^{M} \sum_{l=2}^L \beta_{jl} \mathbf{1}(x_{ij}=l) + \varepsilon_i,
\end{align}  
where $x_{ij}$ is the $j$th feature of the $i$th training observation,
$\varepsilon_i$, $i = 1, \ldots, n_{\text{train}}$, are i.i.d. random variables sampled from the distribution $N(0, 0.01)$, and $\alpha$, $\beta_{jl}$, $j = 1,\ldots,M$, $l = 1,\ldots, L$ are parameters sampled from $N(0,1)$, which are fixed for every experiment. The predictive model, $f(\cdot)$, takes the same form without the noise term (i.e. \eqref{eq:linmod} with $\mathcal{C}_{\text{cont}} = \emptyset$), where the parameters are fit to the $n_{\text{train}}$ training observations using standard linear regression.
Then, we estimate the Shapley values using the different methods from Table \ref{tbl:sim_study_outline2}. 

Table \ref{tbl:sim_study_outline} shows that only ctree and the independence method are used when $M > 4$. This is because for $M > 4$, the methods that require one-hot encoding are too computationally expensive. 
In the same three cases, the number of unique feature combinations ($M^L$) is so large that a subset of the $T=2000$ most likely feature combinations are used instead -- see the discussion related to \eqref{eq:MAE_cat}. 

\begin{table}[ht]
	\centering
	\begin{tabular}{>{\centering}p{0.05\textwidth} >{\centering}p{0.05\textwidth} >{\centering}p{0.05\textwidth} >{\arraybackslash}p{0.35\textwidth} >{\arraybackslash}p{0.25\textwidth}} 
		$M$ & $L$ & $T$ & \textbf{Categorical cut-off values} & \textbf{Methods used}\\
		\midrule 
		3 & 3 & $27$ & $(-\infty, 0, 1, \infty)$ & all\\
		3 & 4 & $81$ & $(-\infty, -0.5, 0, 1, \infty)$ & all\\
		4 & 3 & $64$ & $(-\infty, 0, 1, \infty)$ & all\\
		5 & 6 & $2000$ & $(-\infty, -0.5, -0.25, 0, 0.9, 1, \infty)$ & ctree, independence\\
		7 & 5 & $2000$ & $(-\infty, -0.5, -0.25, 0, 1, \infty)$ & ctree, independence\\
		10 & 4 & $2000$ & $(-\infty, -0.5, 0, 1, \infty)$ & ctree, independence\\ 
		\midrule
	\end{tabular}
	\vspace{0.2cm}
	\caption{An outline of the simulation study when using only categorical features. $M$ denotes the  number of features, $L$ denotes the number of categories, and $T$ denotes the number of unique test observations used to compute \eqref{eq:MAE_cat}.}
	\label{tbl:sim_study_outline}
\end{table}

The results of these experiments are shown in Table \ref{tbl:results}.
When the dependence between the features is small, the performance of each method is almost the same. 
Note that when the correlation, $\rho$, is $0$ (i.e.~the features are independent) and $M\le 4$, the ctree and independence methods perform equally in terms of MAE, and, in fact, give identical Shapley values. This is because when $\rho = 0$, ctree never rejects the hypothesis of independence when fitting any of the trees. As a result, ctree weighs all training observations equally which is analogous to the independence method. This ability to adapt the complexity of the dependence modeling to the actual dependence in the data is a major advantage of the ctree approach.
When $M>4$, the results of the independence and ctree methods for $\rho = 0$ are slightly different. The reason is that when the dimension is large, ctree tests many more hypotheses and therefore is more likely to reject some of the hypotheses. Since the independence method performs better than ctree in these cases, this suggests that the parameter $\alpha$ could be reduced for higher dimensions to improve the performance in low-correlation settings. This remains to be investigated, however.

As expected, the ctree method outperforms the independence method unless the dependence between the features is very small.
The ctree approach also always outperforms (albeit marginally) the empirical, Gaussian, and ctree-onehot approaches.
In addition, a major advantage of using ctree is that it does not require one-hot encoding. Since the computational complexity of computing Shapley values grows exponentially in the number of features (one-hot encoded or not), the computation time for methods requiring one-hot-encoding grows quickly compared to ctree. In Table \ref{tbl:timing}, we show the average run time (in seconds) per test observation of each method. The average is taken over all correlations since the computation times are almost the same for each correlation.  

The empirical method is the fastest amongst the one-hot encoded methods and is still between two and five times slower than the ctree method. This means that if the number of features/categories is large, using one-hot encoding is not suitable. 
For the Gaussian method we calculate \eqref{eq:montecarl} using both 100 and 1000 samples from the conditional distribution. 
Table \ref{tbl:results} shows that the MAE is slightly smaller in the latter case but from Table \ref{tbl:timing}, we see that it is nearly three times slower. 
Such a small performance increase is probably not worth the extra computation time.

\begin{table}[ht]
	\centering
	\begin{tabular}{>{\centering}p{0.05\textwidth} >{\centering}p{0.05\textwidth} >{}p{0.19\textwidth}| >{\centering}p{0.1\textwidth} >{\centering}p{0.1\textwidth} >{\centering}p{0.1\textwidth} >{\centering}p{0.1\textwidth} >{\centering}p{0.1\textwidth} >{\centering\arraybackslash}p{0.1\textwidth}   }   
		\multirow{2}{*}{$M$} & \multirow{2}{*}{$L$} & \multirow{2}{*}{\textbf{Method}} & \multicolumn{6}{c}{$\rho$}\\
		& & & 0 & 0.1 & 0.3 & 0.5 & 0.8 & 0.9 \\ 
		\midrule
		\multirow{6}{*}{3} & \multirow{6}{*}{3} & empirical & 0.0308 & 0.0277 & 0.0358 & 0.0372 & 0.0419 & 0.0430 \\ 
		&  & Gaussian (100)  & 0.0308 & 0.0237 & 0.0355 & 0.0330 & 0.0320 & 0.0384 \\ 
		&  & Gaussian (1000) & 0.0307 & 0.0236 & 0.0354 & 0.0327 & 0.0318 & 0.0383 \\ 
		&  & ctree-onehot & 0.0278 & 0.0196 & 0.0345 & 0.0363 & 0.0431 & 0.0432 \\  
		&  & ctree & \textbf{0.0274} & \textbf{0.0191} & \textbf{0.0302} & \textbf{0.0310} & \textbf{0.0244} & \textbf{0.0259} \\ 
		&  & independence & \textbf{0.0274} & \textbf{0.0191} & 0.0482  & 0.0777 & 0.1546 & 0.2062 \\
		\hline
		\multirow{6}{*}{3} & \multirow{6}{*}{4} & empirical & 0.0491 & 0.0465 & 0.0447  & 0.0639 & 0.0792 & 0.0659 \\ 
		&  & Gaussian (100) &   0.0402 & 0.0350 & 0.0358  & 0.0620 & 0.0762 & 0.0724 \\ 
		&  & Gaussian (1000) &  0.0403 & 0.0353 &   0.0361 & 0.0624 & 0.0763 & 0.0738 \\ 
		&  & ctree-onehot & 0.0324 & 0.0244 & 0.0429  & 0.0617 & 0.0808 & 0.0680 \\ 
		&  & ctree & \textbf{0.0318} & 0.0331 & \textbf{0.0369} & \textbf{0.0422} & \textbf{0.0416} & \textbf{0.0291} \\ 
		&  & independence & \textbf{0.0318} & \textbf{0.0283} &  0.0774  & 0.1244 & 0.2060 & 0.2519 \\ 
		\hline
		\multirow{6}{*}{4} & \multirow{6}{*}{3} & empirical & 0.0385 & 0.0474 &  0.0408 & 0.0502 & 0.0473 & 0.0389 \\ 
		&  & Gaussian (100) &   0.0312 & 0.0381 & 0.0327 & 0.0459 & 0.0475 & 0.0409 \\ 
		&  & Gaussian (1000) &  0.0312 & 0.0385 & 0.0330 & 0.0453 & 0.0480 & 0.0410 \\ 
		&  & ctree-onehot & 0.0234 & 0.0305 & 0.0402 & 0.0530 & 0.0484 & 0.0397 \\ 
		&  & ctree & \textbf{0.0223} & 0.0414 & \textbf{0.0387}  & \textbf{0.0453} & \textbf{0.0329} & \textbf{0.0253} \\ 
		&  & independence &\textbf{0.0223} & \textbf{0.0355} & 0.0961 & 0.1515 & 0.2460 & 0.2848 \\ 
		\hline
		\multirow{2}{*}{5} & \multirow{2}{*}{6} & ctree &  0.0237 & 0.0492 & \textbf{0.0621} & \textbf{0.0760} & \textbf{0.0767} & \textbf{0.0899} \\ 
		&  & independence & \textbf{0.0222} & \textbf{0.0469} & 0.1231 & 0.1803 & 0.2835 & 0.3039 \\ 
		\hline
		\multirow{2}{*}{7} & \multirow{2}{*}{5} & ctree & 0.0209 & \textbf{0.0333} & \textbf{0.0402} & \textbf{0.0542} & \textbf{0.0530} & \textbf{0.0559} \\ 
		&  & independence & \textbf{0.0193} & 0.0345 & 0.0794 & 0.1294 & 0.1908 & 0.2397 \\
		\hline
		\multirow{2}{*}{10} & \multirow{2}{*}{4} & ctree & 0.0169 & \textbf{0.0505} & \textbf{0.0617} & \textbf{0.0607} & \textbf{0.0627} & \textbf{0.0706} \\ 
		&  & independence &  \textbf{0.0153} & 0.0544 & 0.1593 & 0.2180 & 0.3017 & 0.3412 \\ 
		\midrule
	\end{tabular}
\caption{The MAE of each method and correlation, $\rho$, for each experiment. The bolded numbers denote the smallest MAE per experiment and $\rho$.}
\label{tbl:results}
\end{table}

\begin{table}[ht]
	\centering
	\begin{tabular}{>{\centering}p{0.05\textwidth} >{\centering}p{0.05\textwidth} >{\centering}p{0.07\textwidth} >{}p{0.19\textwidth} >{\centering\arraybackslash}p{0.16\textwidth}} 
		$M$ & $L$ & $T$ & \textbf{Method} & \textbf{Mean time per test obs}\\ 
		\midrule
		\multirow{6}{*}{3} & \multirow{6}{*}{3} & \multirow{6}{*}{27} & empirical & 0.086 \\ 
		& & & Gaussian (100) & 4.833 \\ 
		& & & Gaussian (1000) & 13.295 \\ 
		& & & ctree-onehot & 0.338 \\ 
		& & & ctree & 0.040 \\ 
		& & & independence & 0.013 \\ 
		\hline
		\multirow{6}{*}{3} & \multirow{6}{*}{4} & \multirow{6}{*}{64} & empirical & 0.553 \\ 
		& & & Gaussian (100) & 8.041 \\ 
		& & & Gaussian (1000) & 29.160 \\ 
		& & & ctree-onehot & 1.807 \\ 
		& & & ctree & 0.023 \\ 
		& & & independence & 0.007 \\ 
		\hline
		\multirow{6}{*}{4} & \multirow{6}{*}{3} & \multirow{6}{*}{81} & empirical & 0.293 \\ 
		& & & Gaussian (100) & 3.845 \\ 
		& & & Gaussian (1000) & 12.983 \\ 
		& & & ctree-onehot & 0.841 \\ 
		& & & ctree & 0.052 \\ 
		& & & independence & 0.012 \\ 
		\hline
		\multirow{2}{*}{5} & \multirow{2}{*}{6} & \multirow{2}{*}{2000} & ctree & 0.118 \\ 
		& & & independence & 0.030 \\ 
		\hline
		\multirow{2}{*}{7} & \multirow{2}{*}{5} & \multirow{2}{*}{2000} & ctree & 0.590 \\ 
		& & & independence & 0.158 \\
		\hline 
		\multirow{2}{*}{10} & \multirow{2}{*}{4} & \multirow{2}{*}{2000} & ctree & 6.718 \\ 
		& & & independence & 2.066 \\ 
		\midrule
	\end{tabular}
	\caption{The mean run time (in seconds) per test observation, $T$, where the mean is taken over all correlations, $\rho$.}
	\label{tbl:timing}
\end{table}

\subsection{Simulation study with both categorical and continuous features}\label{subsec:both_categorical_and_cont}

We also perform a simulation study with both categorical and continuous features. Because we need to use numerical integration to calculate the true Shapley values (see Section \ref{subsec:calc_true_Shapley}), the computational complexity is large even for lower-dimensional settings. Therefore, we restrict ourselves to an experiment with two categorical features with $L=4$ categories each and two continuous features.
Unless otherwise mentioned, the simulation setup follows that of Section \ref{subsec:only_categorical}. As described in Section \ref{subsec:sim_cat_data}, we simulate dependent categorical/continuous data by only transforming two of the four original Gaussian features. The cut-off vector for the categorical features is set to $(-\infty, -0.5, 0, 1, \infty)$.
Similarly to \eqref{eq:response_mod}, the response is given by
\begin{align*}
y_i = \alpha + \sum_{j=1}^{2} \sum_{l=2}^4 \beta_{jl} \mathbf{1}(x_{ij} = l) + \sum_{j=3}^{4} \gamma_j x_{ij} + \varepsilon_i,
\end{align*}  
for which we fit a linear regression model of the same form without the error term to act as the predictive model $f(\cdot)$.

Then, we estimate the Shapley values using the methods from Table \ref{tbl:sim_study_outline2} except for the Gaussian method with 1000 samples. This method is excluded since Section \ref{subsec:only_categorical} showed that its performance was very similar to that of the Gaussian method with 100 samples but significantly more time consuming.
To compare the performance of the different methods, we sample $T = 500$ observations from the joint distribution of the features and compute the MAE using \eqref{eq:MAE_cat} as described in Section \ref{subsec:evaluation}.

The results are displayed in Table \ref{tbl:results_cont_cat}. 
The Gaussian method is the best performing method when $\rho = 0.1, 0.3, 0.5$ while the empirical and ctree methods are the best performing when $\rho = 0.8$ and $\rho = 0.9$, respectively. The results are not surprising since we only have two categorical features. With more categorical features or categories, we expect that the ctree method would outperform the other ones when $\rho$ is not small.
We also show the run time of each method in Table \ref{tbl:timing_cont_cat}. The one-hot encoded methods are between nine and 75 times slower than the ctree method. This demonstrates, again, the value of using the ctree method when estimating Shapley values with categorical features. 

\begin{table}[ht]
	\centering
	\begin{tabular}{>{\centering}p{0.13\textwidth} >{\centering}p{0.05\textwidth} >{}p{0.19\textwidth}| >{\centering}p{0.1\textwidth} >{\centering}p{0.1\textwidth} >{\centering}p{0.1\textwidth} >{\centering}p{0.1\textwidth} >{}p{0.1\textwidth} >{\centering\arraybackslash}p{0.1\textwidth}   } 
		\multirow{2}{*}{$M$} & \multirow{2}{*}{$L$} & \multirow{2}{*}{\textbf{Method}} & \multicolumn{6}{c}{$\rho$}\\
		& & & 0 & 0.1 & 0.3 & 0.5 & 0.8 & 0.9\\   
		\midrule
		\multirow{6}{*}{2 cont/2 cat} & \multirow{6}{*}{4} & empirical & 0.0853 & 0.0852 & 0.0898 & 0.0913 & \textbf{0.0973} & 0.1027 \\
		& & Gaussian (100) & 0.0570 & \textbf{0.0586} & \textbf{0.0664} & \textbf{0.0662} & 0.1544 & 0.2417 \\ 
	  	& & ctree-onehot & 0.0266 & 0.0714 & 0.1061 & 0.1024 & 0.1221 & 0.1188 \\ 
	  	& & ctree & \textbf{0.0093} & 0.0848 & 0.1073 & 0.1060 & 0.0977 & \textbf{0.0917} \\ 
	  	& & independence & \textbf{0.0093} & 0.0790 & 0.2178 & 0.3520 & 0.5524 & 0.6505 \\ 
	   \midrule
	\end{tabular}
	\caption{The MAE of each method and correlation, $\rho$, for the experiment with two continuous and two categorical features ($L = 4$ categories each). The bolded numbers denote the smallest MAE per $\rho$.}
	\label{tbl:results_cont_cat}
\end{table}

\begin{table}[ht]
	\centering
	\begin{tabular}{>{\centering}p{0.05\textwidth} >{\centering}p{0.05\textwidth} >{\centering}p{0.07\textwidth} >{}p{0.19\textwidth} >{\centering\arraybackslash}p{0.16\textwidth}}  
		$M$ & $L$  &  $T$ & \textbf{Method} & \textbf{Mean time per test obs}\\ 
		\midrule
		\multirow{5}{*}{4} & \multirow{5}{*}{4} & \multirow{5}{*}{500} & empirical & 0.758 \\ 
		 &  &  & Gaussian (100) & 5.914 \\ 
		 &  &  & ctree-onehot & 1.514 \\ 
		 &  &  & ctree & 0.082 \\ 
		 &  &  & independence & 0.057 \\ 
		\midrule
	\end{tabular}
	\caption{The mean run time (in seconds) per test observation, $T$, where the mean is taken over all correlations, $\rho$. The simulation study has two continuous features and two categorical features ($L = 4$ categories each).}
	\label{tbl:timing_cont_cat}
\end{table}

\section{Real data example}\label{sec:data}

Although there is a growing literature of how to explain black-box models, there are very few studies that focus on quantifying the relevance of these methods \cite{adadi2018peeking}. 
This makes it difficult to compare different explainability methods on a real data set since there is no ground truth. 
One partial solution is to compare how different explainability models rank the same features for predictions based on specific test observations.

In this section, we use a data set from the 2018 FICO Explainable Machine Learning Challenge \cite{FICO} aimed at motivating the creation of explainable predictive models. 
The data set is of Home Equity Line of Credit (HELOC) applications made by homeowners.
The response is a binary feature called RiskPerformance that takes the value 1 (`Bad') if the customer is more than 90 days late on his or her payment and 0 (`Good') otherwise. 52 percent of customers have the response `Bad' and 48 percent have the response `Good'. 
There are 23 features: 21 continuous and two categorical (with eight and nine categories, respectively) which can be used to model the probability of being a `Bad' customer. 
Features with the value -9 are assumed missing. We remove the rows where all features are missing.

We first use this data set to compare the Shapley values calculated using the independence approach with those calculated with our ctree approach. Then, for a few test observations, we see how the Shapley explanations compare with the explanations from the 2018 FICO challenge Recognition Award winning team from Duke University \cite{chen2018interpretable} (hereafter referred to as just `Duke'). 

After removing the missing data and a test set of 100 observations, we use the remaining 9,765 observations to train a 5-fold cross validation (CV) model using xgboost \cite{Chen2016} and then average these to form the final model.
Our model achieves an accuracy of 0.737 (compared to Duke's accuracy of 0.74).
In our experience, explanation methods often behave differently when there is dependence between the features.
As a measure of dependence, we use the standard Pearson correlation for all continuous features, Cramer’s V correlation measure \cite{cramir1946mathematical} for categorical features, and the correlation ratio \cite{fisher1992statistical} for continuous and categorical features.
The feature `MSinceMostRecentInqexcl7days' is the least correlated with the rest of the features with correlations between -0.109 and 0.07. The 22 other features are strongly correlated with at least one other feature (max absolute correlations between 0.4 and 0.99). 

\begin{figure}[ht]
	\centering
	\includegraphics[width = 10cm, height = 10cm]{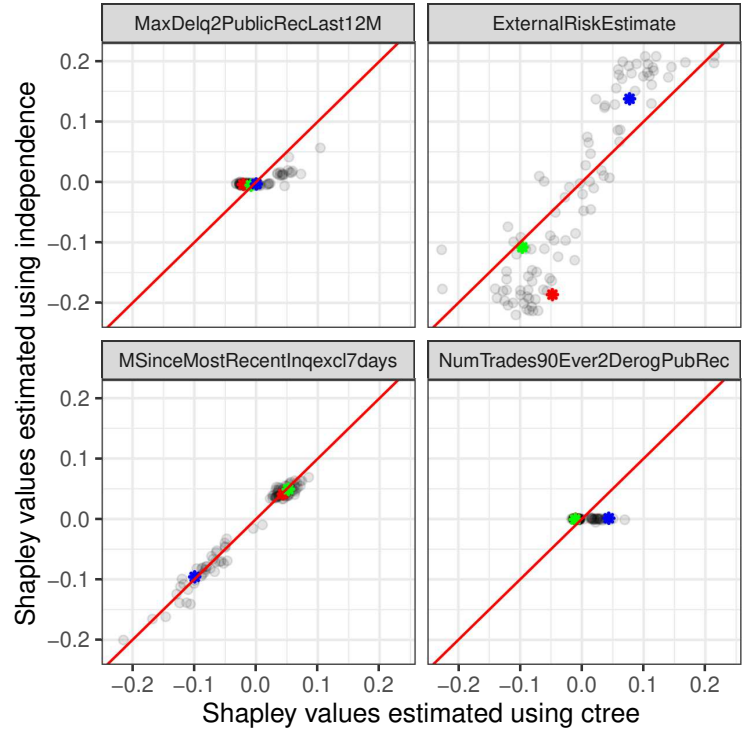}
	\caption{The Shapley values of 100 test observations calculated using both the independence and the ctree method for four of the 23 features.}
	\label{fig:four_features}
\end{figure}

Turning to the Shapley value comparisons, we estimate the Shapley values of the features belonging to the 100 test observations using the independence approach and the ctree approach. Then, we plot the Shapley value estimates against each other for a selection of four features in Figure \ref{fig:four_features}.
The top left panel shows the Shapley values of one of the two categorical features (`MaxDelq2PublicRecLast12M'). 
Both methods give Shapley values fairly close to 0 for this feature, but there are some differences.
The top right panel shows an example of a feature (`ExternalRiskEstimate') where the two methods estimate quite different Shapley values.
Since these are some of the largest (absolute) Shapley values, this is one of the most influential features. We also see that for most test observations, the independence method estimates more extreme Shapley values than the ctree method.

The bottom left panel shows a feature (`MSinceMostRecentInqexcl7days') where the two methods estimate relatively similar Shapley values. As noted above, this feature is the least correlated with the rest of the features. We believe the two methods behave similarly because the independence method performs best when dealing with nearly independent features. 
Finally, the bottom right panel shows a feature (`NumTrades90Ever2DerogPubRec') where the independence method assigns most test observations a Shapley value very close to 0 while the ctree method does not. Although not plotted, we see this trend for 6 out of the 23 features. We notice that each of these 6 features are highly correlated with at least one other feature (max absolute correlation between 0.46 and 0.99). We suspect that the methods behave differently because of the independence method's failure to account for dependence between features.
We also colour three random test observations to show that for some test observations (say `green'), 
all Shapley values are estimated very similarly for the two methods, while for others (say `blue'), the methods are sometimes quite different.

We also attempt to compare explanations based on Shapley values (estimated with either the independence or ctree method) with those based on Duke's approach \cite{chen2018interpretable}
We reinforce that there is \textit{no ground truth} when it comes to explanations and that Duke does not use Shapley values in their solution. Therefore, we only compare how these three methods \textit{rank} feature importance for specific test observations.  

Duke's explainability approach is based on 10 smaller regression models fit to 10 partitions (below called `groups') of the features. 
They use a combination of learned weights and risks to calculate the most influential group for each customer/test observation. 
Based on our understanding, if the customer has a large predicted probability of being `Bad', the largest $\text{weight} \times \text{risk}$ is the most influential group (given rank 1), while for small predicted probabilities, the largest weight divided by risk is given rank 1. It is not clear how they rank medium-range predictions.

We speculate that Duke does not properly account for feature dependence since they fit 10 independent models that only interact using an overall learned risk for each model. 
To test this hypothesis, we compare the group rankings of a few\footnote{Duke's explanations were not readily available. For a given test observation, we had to manually input 23 feature values into a web application to get an explanation. Therefore, it was too time consuming to compare many test observations.} customers/test observations calculated by 1. Duke, 2. The Shapley approach under the independence assumption, and 3. Our new Shapley approach that uses conditional inference trees.  

To calculate group importance based on Shapley values for a given test observation, we first estimate the Shapley values of each feature either either the independence or ctree approach. Then, we sum the Shapley values of the features belonging to the same group. This gives 10 new grouped Shapley values. Finally, we rank the grouped Shapley values by giving rank 1 to the group with the largest absolute grouped Shapley value.  
While the prediction models being compared here are different (we use an xgboost model while Duke does not), the two models have very similar overall performance and give similar predictions to specific test observations. We believe this validates the rough comparison below.

We observe that for test observations with a large predicted probability of being `Bad', there is little pattern among the rankings calculated by the three explanation methods. However, when Duke and independence give a group the same ranking out of 10 (and ctree does not), we notice that this group includes at least one feature that is very correlated with a feature in another group.  
On the other hand, for test observations with a small predicted probability of being `Bad', we notice that all three explanation methods rank the groups similarly. Again, the only time ctree ranks a group differently than Duke and independence (but Duke and independence rank similarly), the group includes at least one feature highly correlated with a feature in another group. 

\section{Conclusion}\label{sec:conclusion}

The aim of this paper was to extend \cite{lundberg2018consistent} and \cite{aas2019Explaining}'s Shapley methodology to explain mixed dependent features using conditional inference trees \cite{hothorn2006unbiased}. We showed in two simulation studies that when the features are even mildly dependent, it is advantageous to use our ctree method over the traditional independence method. Although ctree often has comparable accuracy to some of \cite{aas2019Explaining}'s methods, those methods require transforming the categorical features to one-hot encoded features. 
We demonstrated that such one-hot encoding leads to a substantial increase in computation time, making it infeasible in high dimensions.

We also demonstrated our methodology on a real financial data set. 
We first compared the Shapley values of 100 test observations calculated using the independence and ctree approaches. We noticed that the methods performed similarly for an almost independent feature but otherwise performed quite differently. 

Then, we compared explanations based on the independence/ctree Shapley approaches with those based on Duke's approach \cite{chen2018interpretable}. We had to fall back to comparing how the different methods ranked features rather than the explanations themselves because there is no ground truth when it comes to explainability.
It was difficult to argue for one explanatory approach over another; however, Duke's rankings seemed to agree more with the rankings based on Shapley values calculated under independence (which we saw was inaccurate in simulation studies), than with our proposed ctree based Shapley value estimation method.

\bibliographystyle{splncs04}
\bibliography{references}

\end{document}